\documentclass[10pt,letterpaper]{article}
\usepackage[margin=1in]{geometry}

\usepackage{amsmath,amssymb}

\usepackage{changepage}

\usepackage{textcomp,marvosym}

\usepackage{cite}

\usepackage{nameref,hyperref}

\usepackage[nopatch=eqnum]{microtype}
\DisableLigatures[f]{encoding = *, family = * }

\usepackage[table]{xcolor}

\usepackage{array}

\newcolumntype{+}{!{\vrule width 2pt}}

\newlength\savedwidth

\raggedright
\usepackage[aboveskip=1pt,labelfont=bf,labelsep=period,justification=raggedright,singlelinecheck=off]{caption}

\makeatletter
\renewcommand{\@biblabel}[1]{\quad#1.}
\makeatother

\usepackage{graphicx}
\begin{document}
\vspace*{0.2in}

\begin{flushleft}
{\Large
\textbf{Predicting household water consumption using satellite and street view images in two Indian cities} % Please use "sentence case" for title and headings (capitalize only the first word in a title (or heading), the first word in a subtitle (or subheading), and any proper nouns).
}
\newline
\\
Qiao Wang\textsuperscript{1*\ddag},
Joseph George\textsuperscript{2\ddag},
% Name3 Surname\textsuperscript{2,3\textcurrency},
% Name4 Surname\textsuperscript{2},
% Name5 Surname\textsuperscript{2\ddag},
% Name6 Surname\textsuperscript{2\ddag},
% Name7 Surname\textsuperscript{1,2,3*},
% with the Lorem Ipsum Consortium\textsuperscript{\textpilcrow}
\\
\bigskip
\textbf{1} World Bank, Washington, District of Columbia, USA
\\
\textbf{2} Thomas Jefferson High School for Science and Technology, Alexandria, Virginia, USA
\\
% \textbf{3} Affiliation Dept/Program/Center, Institution Name, City, State, Country\\
\bigskip

% Insert additional author notes using the symbols described below. Insert symbol callouts after author names as necessary.
% 
% Remove or comment out the author notes below if they aren't used.
%
% Primary Equal Contribution Note
% \Yinyang These authors contributed equally to this work.

% Additional Equal Contribution Note
% Also use this double-dagger symbol for special authorship notes, such as senior authorship.
\ddag These authors also contributed equally to this work.

% \textcurrency Current Address: Dept/Program/Center, Institution Name, City, State, Country % change symbol to "\textcurrency a" if more than one current address note
% \textcurrency b Insert second current address 
% \textcurrency c Insert third current address

% \dag Deceased

% Group/Consortium Author Note
% \textpilcrow Membership list can be found in the Acknowledgments section.

% Use the asterisk to denote corresponding authorship and provide email address in note below.
* qwang3@worldbank.org

\end{flushleft}

% For PLOS Medicine research article authors, please structure your abstract
% with "Background", "Method and Findings" and "Conclusion" sections per
% journal requirements.

% For PLOS Neglected Tropical Diseases research article authors, please
% structure your abstract with "Background", "Methodology", "Findings", and
% "Conclusion" sections per journal requirements.
%
\section*{Abstract}
Monitoring household water use in rapidly urbanizing regions is hampered by costly, time-intensive enumeration methods and surveys. We investigate whether publicly available imagery—satellite tiles, Google Street View (GSV) segmentation—and simple geospatial covariates (nightlight intensity, population density) can be utilized to predict household water consumption in Hubballi-Dharwad, India. We compare four approaches: survey features (benchmark), image embeddings (satellite, GSV, combined), and GSV semantic maps with auxiliary data. Under an ordinal classification framework, GSV segmentation plus remote-sensing covariates achieves 0.55 accuracy for water use, approaching survey-based models (0.59 accuracy). Error analysis shows high precision at extremes of the household water consumption distribution, but confusion among middle classes is due to overlapping visual proxies. We also compare and contrast our estimates for household water consumption to that of household subjective income. Our findings demonstrate that open-access imagery, coupled with minimal geospatial data, offers a  promising alternative to  obtaining reliable household water consumption estimates using surveys in urban analytics.

%\linenumbers

% Use "Eq" instead of "Equation" for equation citations.
\section*{Introduction}
Cities across the world are urbanizing at unprecedented speed, offering opportunities for economic growth but also amplifying challenges of equitable service delivery. Harnessing the advantages of urbanization requires timely, fine-grained understanding of socio-economic conditions, yet traditional surveys remain costly, labor-intensive, and limited in spatial coverage. In India, where urban growth is especially rapid, ensuring fair access to essential service, particularly domestic water, has become a pressing policy concern. Recent advances in computer vision and remote sensing provide new opportunities to generate scalable, high-resolution socio-economic indicators from ubiquitous visual data such as satellite imagery and Google Street View (GSV), reducing reliance on conventional approaches such as complete enumeration and estimations using household surveys.

In this study, we evaluate the effectiveness of publicly available visual and geospatial data—including satellite imagery, GSV-based semantic segmentation, nighttime lights, and population density—for predicting household water consumption in Hubballi-Dharwad, India and compare it with predictions for subjective household income. Prediction using survey-based features are used as a performance benchmark. The most effective model incorporates semantic features extracted from GSV segmentation maps along with publicly available remote sensor data such as nightlight intensity and population density. This combination consistently outperforms other image-only models across both household water consumption and household subjective income tasks and approaches the performance of survey-based features. This suggests a promising direction for scalable, image-driven urban analytics that reduce reliance on expensive and time-consuming traditional data collection.

Accurate, fine-grained estimates of household water use and household income are essential for equitable urban planning, yet traditional household surveys are costly, time-consuming, and lack spatial representativeness to identify the most vulnerable. Nighttime lights (NTL) at $1\,\text{km}^2$ capture national GDP trends but do not resolve variation below $\$1.90$/day (low uniform luminosity)~\cite{engstromPovertySpaceUsing2022}. However, mobile phone data, though rich in behavioral signals, face scalability and privacy barriers in diverse regions~\cite{jeanCombiningSatelliteImagery2016}.

Satellite imagery has been applied extensively for socio‐economic and environmental measurement tasks: estimating poverty and asset‐wealth at survey cluster scales~\cite{jeanCombiningSatelliteImagery2016}, ~\cite{yehUsingPubliclyAvailable2020} or small-area administrative units~\cite{engstromPovertySpaceUsing2022}; multi‐task prediction of developmental proxies—roof material, lighting source, and water access—prior to income classification~\cite{pandeyMultitaskDeepLearning2018}; and building-footprint segmentation to predict neighborhood‐level water consumption~\cite{mohantyUnderstandingUrbanWater2022}.  Most of these approaches rely on a single satellite data source and employ Convolutional Neural Networks (CNNs) transfer learning to extract features for downstream regression or classification.

Street‐level imagery has been harnessed to infer neighborhood‐scale socio‐economic and public‐health outcomes across diverse settings: vehicle cues predict income and voting patterns in the U.S.~\cite{gebruUsingDeepLearning2017}; saliency‐map analyses classify income brackets via greenery and concrete cues in Oakland~\cite{acharyaNeighborhoodWatchUsing2017}; CNNs estimate income, health, and crime deciles in U.K. cities~\cite{suelMeasuringSocialEnvironmental2019}; panoptic segmentation of community‐sourced imagery predicts poverty, population, and BMI in India and Kenya~\cite{leePredictingLivelihoodIndicators2021}; and semantic quantification of urban features explains travel behavior and poverty rates in the U.S.~\cite{fanUrbanVisualIntelligence2023}. These studies predominantly employ CNNs‐based object detection or segmentation to derive visual features, often augmented by transfer learning, fine-tuning, and postprocessing techniques.

Recent work has demonstrated the power of jointly modeling aerial and street-level imagery to capture both macro-scale spatial context and human-scale visual cues. Such architectures typically combined convolutional feature extractors for, enabling end-to-end training to predict building energy efficiency~\cite{mayerEstimatingBuildingEnergy2023}. Similarly, hedonic pricing studies augment conventional housing attributes with housing visual features extracted from aerial and street images using pre-trained networks for property valuation~\cite{lawTakeLookUsing2019}. The most novel approaches employ dual-branch architectures: one processing multi‐angle panoramas to capture street‐level context and the other encoding high-resolution satellite patches—and fuse their features end-to-end, leveraging complementary perspectives for more accurate urban predictions~\cite{suelMultimodalDeepLearning2021}.

Building on these insights, our work is the first to systematically compare survey‐based models, single‐modality embeddings, joint fusion, and GSV semantic segmentation with geospatial covariates for predicting income and water consumption in Hubballi‐Dharwad, India—offering a comprehensive evaluation of accuracy, scalability, and practical trade‐offs for urban socio‐economic inference.

\section*{Materials and methods}
\subsection*{Data}
\subsubsection*{Customer survey data}

The customer survey data from Hubballi-Dharwad, Karnataka, India, provides comprehensive insights into domestic, commercial, and industrial water customers. Conducted via door-to-door interviews in 2022, the survey documented over 200,000 customers, capturing socio-economic characteristics, property attributes, and water usage behaviors. The dataset includes roughly 90,000 domestic customers and records property owner information, family income, caste, property type (e.g., house, apartment), roof construction, built-up area, water connection details, storage methods, sewage connection type, and rainwater harvesting implementation. A geo-referenced base map created using satellite imagery was used to link customer locations with the network map, enabling unique indexing of each customer based on geo-coded data.  

\subsubsection*{Water billing data}

The water billing data, collected exclusively in the continuous water service area (24/7 supply), provides detailed records of water usage, billing status, and meter readings for domestic customers. After preprocessing, the dataset included more than 30,000 domestic customers, with water billing information available for 24,366 customers. Additionally, domestic properties with self-reported monthly income totaled 34,811 after restricting the analysis to the continuous water supply service area.

To ensure data quality, only records from fully functional meters were retained. These records, spanning January to May 2023, were aggregated to compute average monthly water consumption for each customer. Using unique Revenue Register Numbers (RRNo), the water billing data was integrated with the customer survey dataset, enabling a comprehensive analysis of water usage patterns in relation to socio-economic characteristics. 

\begin{figure}
    \centering
    \includegraphics[width=0.5\linewidth]{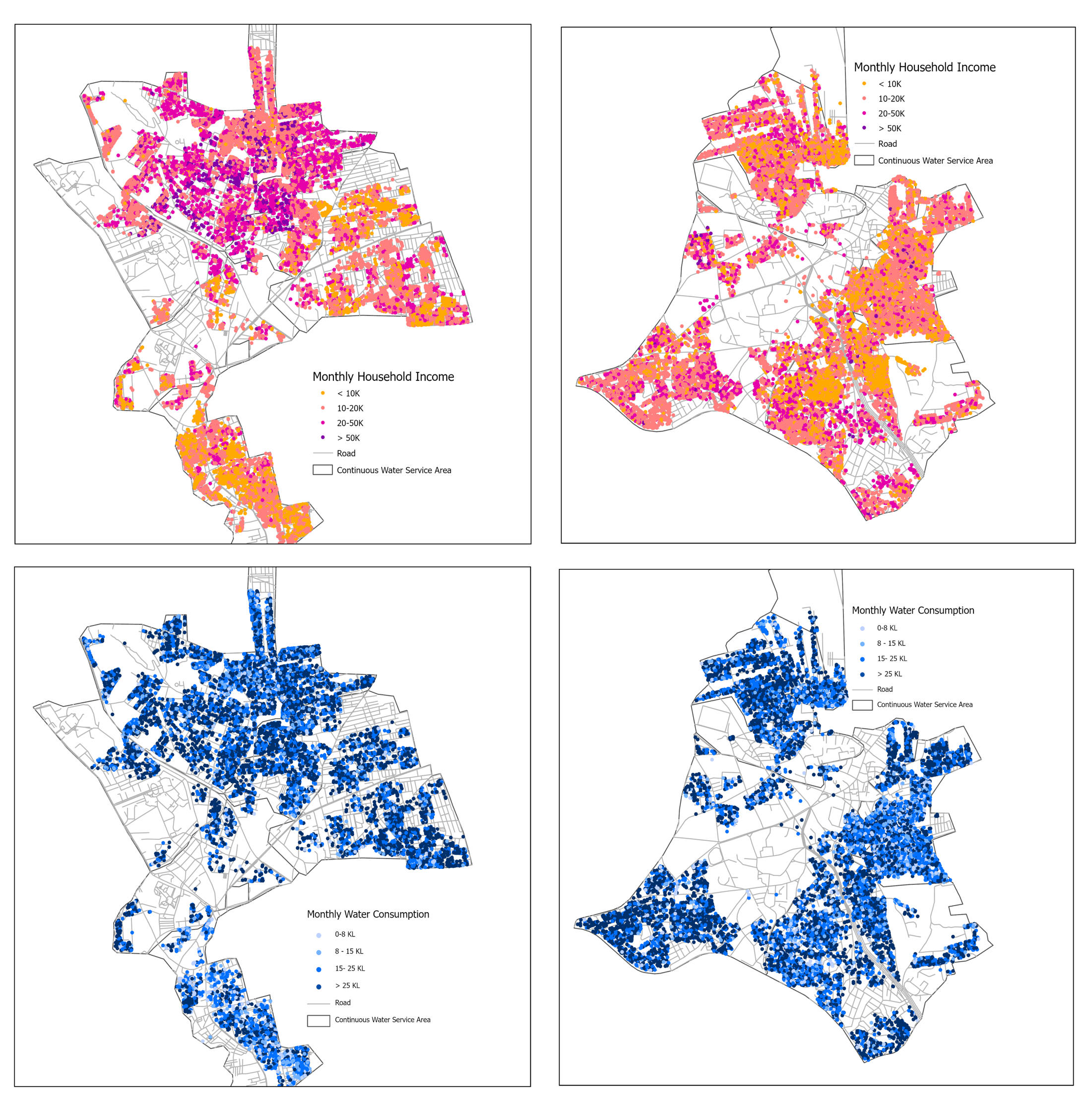}
    \caption{\textbf{Spatial distribution of monthly household income and water consumption.}
Self-reported monthly income (top) and average monthly water consumption (bottom) are shown for continuous water supply zones in Dharwad (left) and Hubballi (right). Maps were produced using Esri’s ArcGIS Pro 3.1.4.}
    \label{fig1}
\end{figure}

Fig~\ref{fig1} illustrates the spatial distribution of monthly household income and average monthly water consumption across the continuous water supply zones in Hubballi-Dharwad. Household income is self-reported and categorized into four ordinal brackets:Rs.~0--10K, Rs.~10--20K, Rs.~20--50K, and $>$Rs.~50K, with the Rs.~10--20K group comprising the largest share (18,982 households). Water consumption levels are grouped based on the municipal tariff structure into four consumption tiers: 0--8, 8--15, 15--25, and $>$25~kiloliters per month.

% For figure citations, please use "Fig" instead of "Figure".
% Place figure captions after the first paragraph in which they are cited.

\subsubsection*{Satellite imagery and street view images}

Satellite and Google Street View (GSV) images were collected to capture macro- and micro-urban characteristics at each customer location in Hubballi-Dharwad. Geo-coordinates from the customer survey dataset were used to automate the download process via the Google Maps API.

Satellite images were retrieved using the Google Static Maps API, with a zoom level of 30 to capture detailed urban features, including building density, greenery, and road networks. Images were downloaded at a resolution of 640x640 pixels, balancing visual clarity with processing efficiency. A total of approximately 30,000 satellite images were collected.

Street view images were obtained using the Google Street View API. Each image was captured at a resolution of 600x400 pixels with a 90° field of view (FOV), providing detailed information on streetscape elements such as building facades, road conditions, and vegetation. A total of approximately 28,000 street view images were collected. Both image types were cropped to remove extraneous artifacts and standardized to RGB format for downstream processing. 

Auxiliary geospatial data were incorporated to complement the satellite and street view imagery, providing additional insights into urban characteristics and their influence on water consumption and income prediction. All geospatial datasets were retrieved using geo-coordinates from the customer survey, ensuring precise spatial alignment with the study area.

Nightlight luminosity~\cite{elvidgeAnnualTimeSeries2021}  and population density~\cite{sorichettaHighresolutionGriddedPopulation2015} were derived from Google Earth Engine, serving as proxies for urban development and population distribution at a granular level. Building footprints, detailing the spatial layout of structures, were extracted from OpenStreetMap. Additionally, building heights were obtained from the Open Buildings 2.5D Temporal Dataset~\cite{sirkoHighresolutionBuildingRoad2024}, enabling the analysis of vertical urban density and its correlation with resource consumption. Together, these datasets enriched the predictive modeling framework by integrating diverse dimensions of urban morphology.

\subsection*{Methods}

All analyses were conducted in Python (version 3.10) using open-source libraries, including PyTorch for deep learning and semantic segmentation, scikit-learn and LightGBM for machine learning models.

\subsubsection*{Survey-based machine learning}

We first trained machine learning models using household-level features collected from the customer survey, encompassing socio-economic status (including household size, number of residents, and caste affiliation), housing characteristics (capturing property type, subtype, roof material, and number of floors), and water infrastructure (such as connection size, supply frequency and duration, perceived water pressure, water storage methods, and whether the household had a sewage connection or implemented rainwater harvesting).

To accommodate the ordinal nature of both water consumption and income brackets, we implemented an ordinal classification framework. The ordinal loss function decomposes the multiclass problem into a sequence of binary subproblems, each predicting whether an observation belongs to a class higher than a given threshold. Class probabilities are reconstructed from these conditional classifiers, enabling consistent treatment of class order. We evaluated four widely used models: logistic regression, which provides a strong linear baseline; random forest, a non-parametric ensemble method effective for structured tabular data; gradient boosting, which builds successive learners to correct residual errors; and LightGBM, a gradient boosting framework optimized for speed and memory efficiency. Each model was trained with five-fold cross-validation, and performance was assessed using the multiclass accuracy and a receiver operating characteristic–area under the curve (ROC-AUC) score to account for the ordinal structure of the target variables.

\subsubsection*{Satellite and GSV image embeddings from transfer learning}

We utilized a convolutional neural network based on EfficientNet-B0, pre-trained on ImageNet, to extract compact feature embeddings from both satellite and Google Street View (GSV) images. The model retains the full convolutional backbone of EfficientNet-B0, which includes multiple mobile inverted bottleneck convolution (MBConv) blocks with depthwise separable convolutions and squeeze-and-excitation operations for improved efficiency and accuracy. After the final convolutional layer, we applied global average pooling to reduce each feature map to a single value, effectively summarizing spatial information across the image. The resulting feature vector was then flattened into a one-dimensional array and passed through a fully connected linear layer to reduce its dimensionality from 1280 to 256.

The entire network was used in inference mode, without fine-tuning any of the pre-trained weights. For each image, this pipeline produced a 256-dimensional embedding capturing high-level visual semantics. These embeddings were extracted for all satellite and street-level images and subsequently used as input features in traditional machine learning models for multi-class ordinal classification of household income and water consumption levels.

\subsubsection*{Semantic segmentation}

\begin{figure}
    \centering
    \includegraphics[width=1\linewidth]{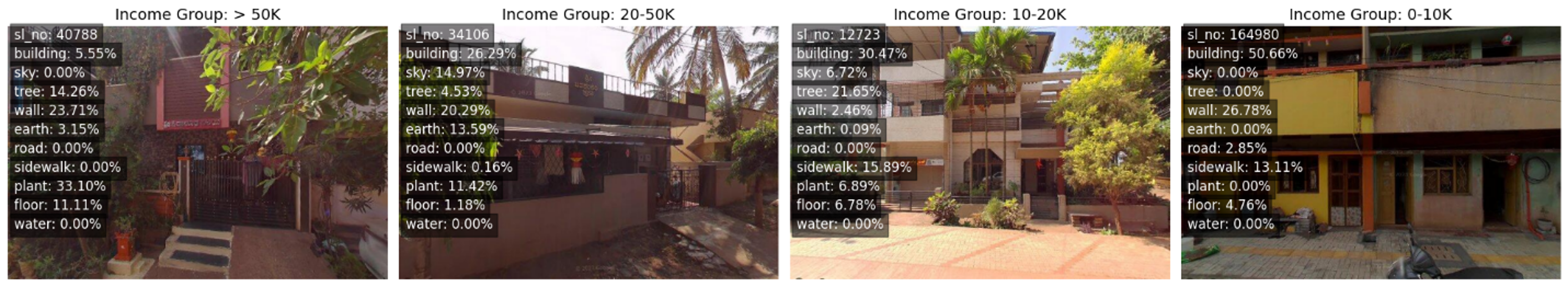}
    \caption{\textbf{Illustrative examples street view images with semantic segmentation features.}
Pixel-level percentages for each detected feature (e.g., buildings, walls, sidewalks, vegetation, and sky) are shown on the left side of each image.}
    \label{fig2}
\end{figure}

To capture fine-grained urban form features from the built environment, we applied semantic segmentation to the Google Street View (GSV) images using a pre-trained model from the MIT ADE20K dataset~\cite{zhouSemanticUnderstandingScenes2019}.  The model architecture consists of a ResNet50-dilated encoder that preserves spatial resolution while extracting deep semantic features, paired with a Pyramid Pooling Module (PPM) decoder with deep supervision for precise pixel-level classification. This setup enables classification of 150 object categories commonly found in urban settings, including roads, buildings, vegetation, sidewalks, vehicles, and sky. Using PyTorch, we applied the model to each GSV image and recorded the proportion of pixels corresponding to each semantic class, effectively summarizing the streetscape composition around each household location (See exampls in Fig~\ref{fig2}) .

\subsubsection*{Address data imbalance using over-sampling}

To mitigate class imbalance in the training data—particularly underrepresentation of households in the highest and lowest income or consumption brackets—we employed the Synthetic Minority Over-sampling Technique (SMOTE). SMOTE  generates synthetic samples by interpolating between existing minority class instances, effectively augmenting the training dataset without duplication~\cite{chawlaSMOTESyntheticMinority2002}. This method was applied independently to the extracted features from satellite and street view images prior to model training. 

% Results and Discussion can be combined.
\section*{Results}
\subsection*{Model Performance}

\begin{figure}
    \centering
    \includegraphics[width=0.95\linewidth]{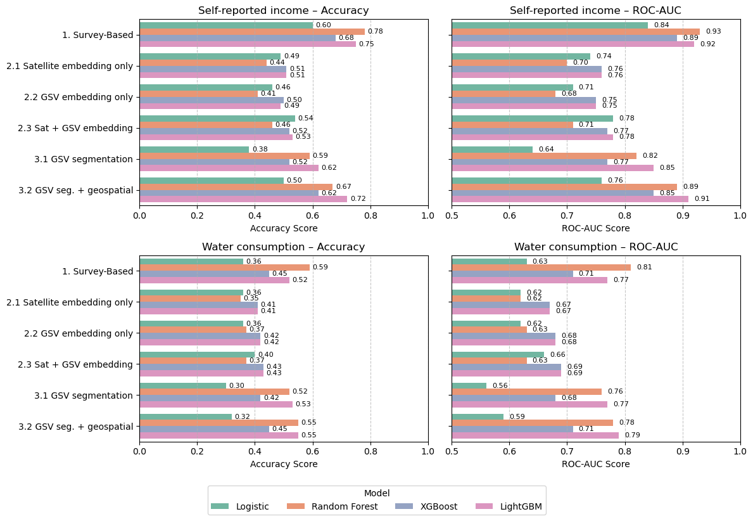}
    \caption{\textbf{Comparative performance of models predicting household income and water consumption using survey, image, and geospatial features.}
Each bar shows Accuracy or ROC-AUC score across four classifiers (Logistic Regression, Random Forest, XGBoost, and LightGBM) under six feature settings, including survey-based, image embeddings, semantic segmentation, and geospatial augmentation.}
    \label{fig3}
\end{figure}

The results demonstrate the feasibility of leveraging publicly available visual and geospatial data, such as satellite imagery and Google Street View (GSV), to predict urban household water consumption and income. While models trained on survey-based features achieve the highest predictive performance, particularly with Random Forest and LightGBM classifier (validation accuracy of 0.52-0.59 for water consumption, and 0.75-0.78 for income), these models rely on high-cost, labor-intensive data collection processes that are inherently difficult to scale. The comparative results across all feature sets and models are summarized in Fig~\ref{fig3}.

In contrast, our experiments show that transfer learning-based image embeddings and GSV semantic segmentation offer scalable alternatives that approximate survey-level performance, especially when combined with contextual geospatial covariates such as population density, nighttime lights, DMA zoning, and building square footage. Among these alternative modalities, the integration of GSV segmentation with geospatial data yields the most consistent and robust results. For income prediction, lightGBM achieves an accuracy of 0.72 and ROC-AUC of 0.91 for using this combined feature set—closely matching survey-based benchmarks. Similarly, for water consumption, this configuration attains competitive performance (accuracy 0.55, ROC-AUC 0.79). These findings underscore the potential of open-access street-level imagery and remote sensing as viable proxies for household-level socioeconomic indicators in urban environments. Based on its superior generalization and stability, the GSV segmentation plus geospatial feature set with lightGBM model will be selected for further optimization via hyperparameter tuning in subsequent experiments.

\subsection*{Hyperparameter Tuning and Feature importance}

To further improve model generalization and robustness, we performed an extensive grid search to optimize the LightGBM classifier within the ordinal classification framework. The search space spanned key hyperparameters including the number of estimators, tree depth, learning rate, leaf complexity, child sample size, and column and row sampling rates. A 5-fold cross-validation was employed with a dual-objective evaluation strategy, using both accuracy and ROC-AUC as metrics. The model achieving the highest validation accuracy (0.72 for income prediction and 0.61 for water consumption prediction) was selected as the optimal configuration. 

\begin{figure}
    \centering
    \includegraphics[width=1\linewidth]{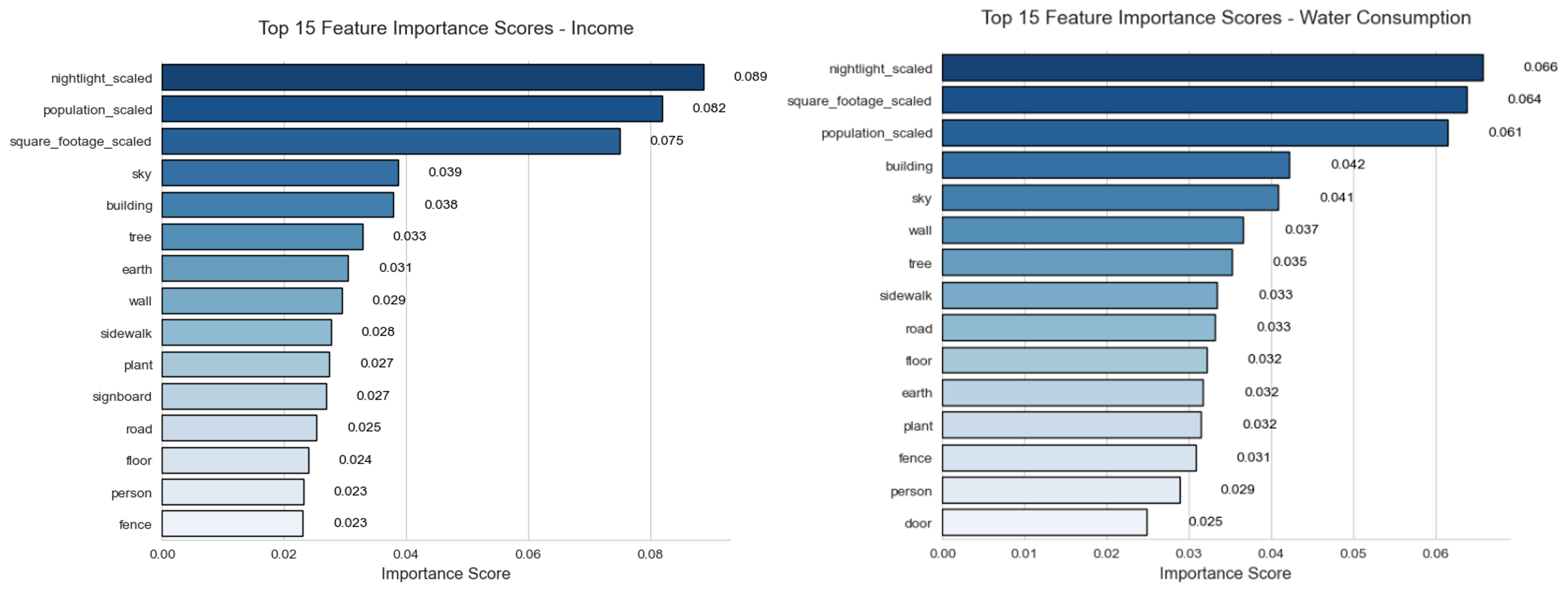}
    \caption{\textbf{Feature importance scores from the best-performing LightGBM model predicting household income and water consumption levels}
Left: Top 15 feature importance scores to predict income. Right: Top 15 feature importance scores to predict water consumption.}
    \label{fig4}
\end{figure}

Following model selection, we conducted a feature attribution analysis to interpret model decisions and uncover the most influential predictors. As illustrated in Fig~\ref{fig4}, the ranked importance scores reveal consistent dominance of geospatial covariates, complemented by informative visual features extracted from GSV segmentation. Specifically, nighttime lights, population density, and building square footage emerged as the top three contributors. Notably, semantic segmentation outputs from GSV images, such as the proportions of sky, building, wall, and vegetation, also demonstrated high importance, highlighting the effectiveness of visual context in approximating socioeconomic conditions. The aligned patterns across income and water consumption models reinforce the role of urban morphology and infrastructure signals captured through open-access imagery and remote sensing in predictive modeling. These findings support the feasibility of building scalable, interpretable, and high-performing socioeconomic models without relying on extensive household surveys.

\subsection*{Error analysis}

\begin{figure}
    \centering
    \includegraphics[width=1\linewidth]{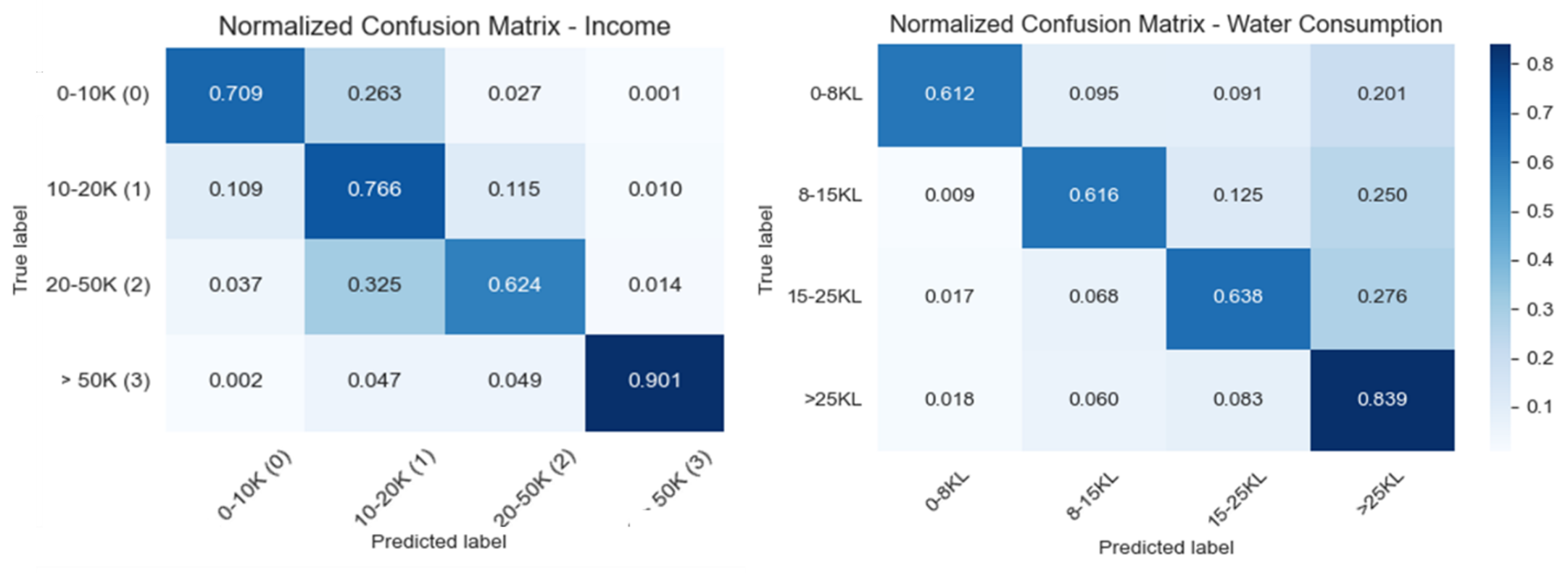}
    \caption{\textbf{Confusion matrix of the best performing model on the validation set.}
Left: Normalized confusion matrix -- income. Right: Normalized confusion matrix -- water consumption.}
    \label{fig5}
\end{figure}

Using normalized confusion matrices (Fig.~\ref{fig5}), we observe that the model performs best at the extremes of both income and water consumption distributions. For income, 90.1 percent of households in the $>$Rs.~50K group and 70.9 percent in the Rs.~0--10K group are correctly classified, while middle-income groups show greater confusion—particularly Rs.~20--50K, which is often misclassified as Rs.~10--20K. A similar pattern appears in water consumption: the model correctly predicts 83.9 percent of $>$25KL cases, but accuracy drops for the lowest group (0--8KL, 61.2 percent) with frequent misclassification into higher categories. These patterns suggest that the model captures extreme cases well but struggles to differentiate overlapping feature patterns in middle-range groups.

\begin{figure}
    \centering
    \includegraphics[width=1\linewidth]{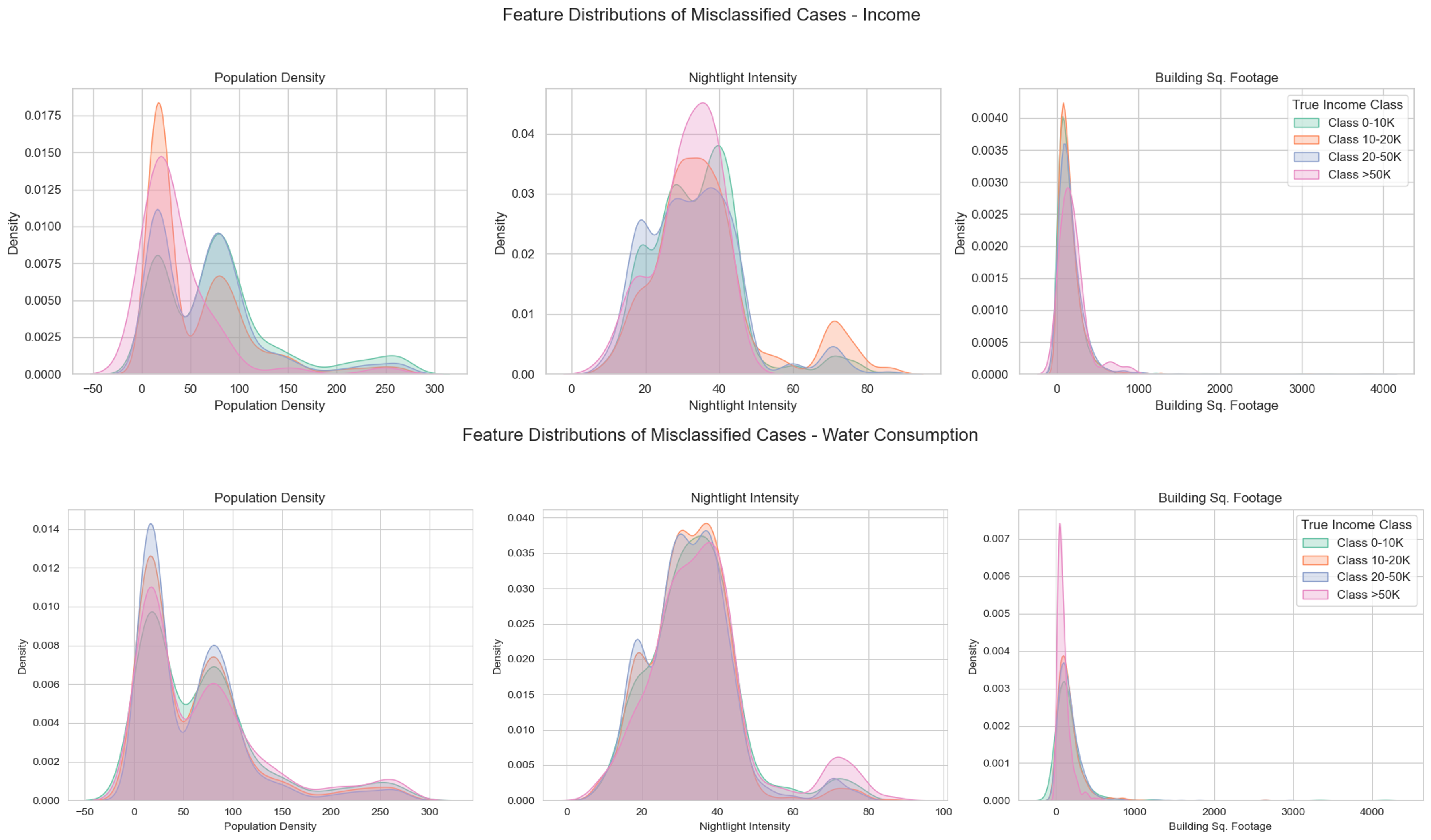}
    \caption{\textbf{Feature distributions of misclassified cases for income and water consumption predictions.}
Top panel: Feature distributions of misclassified cases -- income. Bottom panel: Feature distributions of misclassified cases -- water consumption.}
    \label{fig6}
\end{figure}

To better understand model limitations, we examined feature distributions for misclassified cases in both income and water consumption predictions. The KDE in plots in Fig.~\ref{fig6} reveal substantial overlap in population density, nightlight intensity, and building square footage across income classes, especially within the middle-income (10–50K) and mid-consumption (8–25KL) ranges. This overlap suggests that these features, while predictive in aggregate, lack sufficient discriminatory power at individual levels in boundary cases. For instance, some low-income households misclassified as higher-income exhibit unexpectedly high nightlight intensity or larger built-up area, potentially due to mixed land use or proximity to commercial structures.

\begin{figure}
    \centering
    \includegraphics[width=0.75\linewidth]{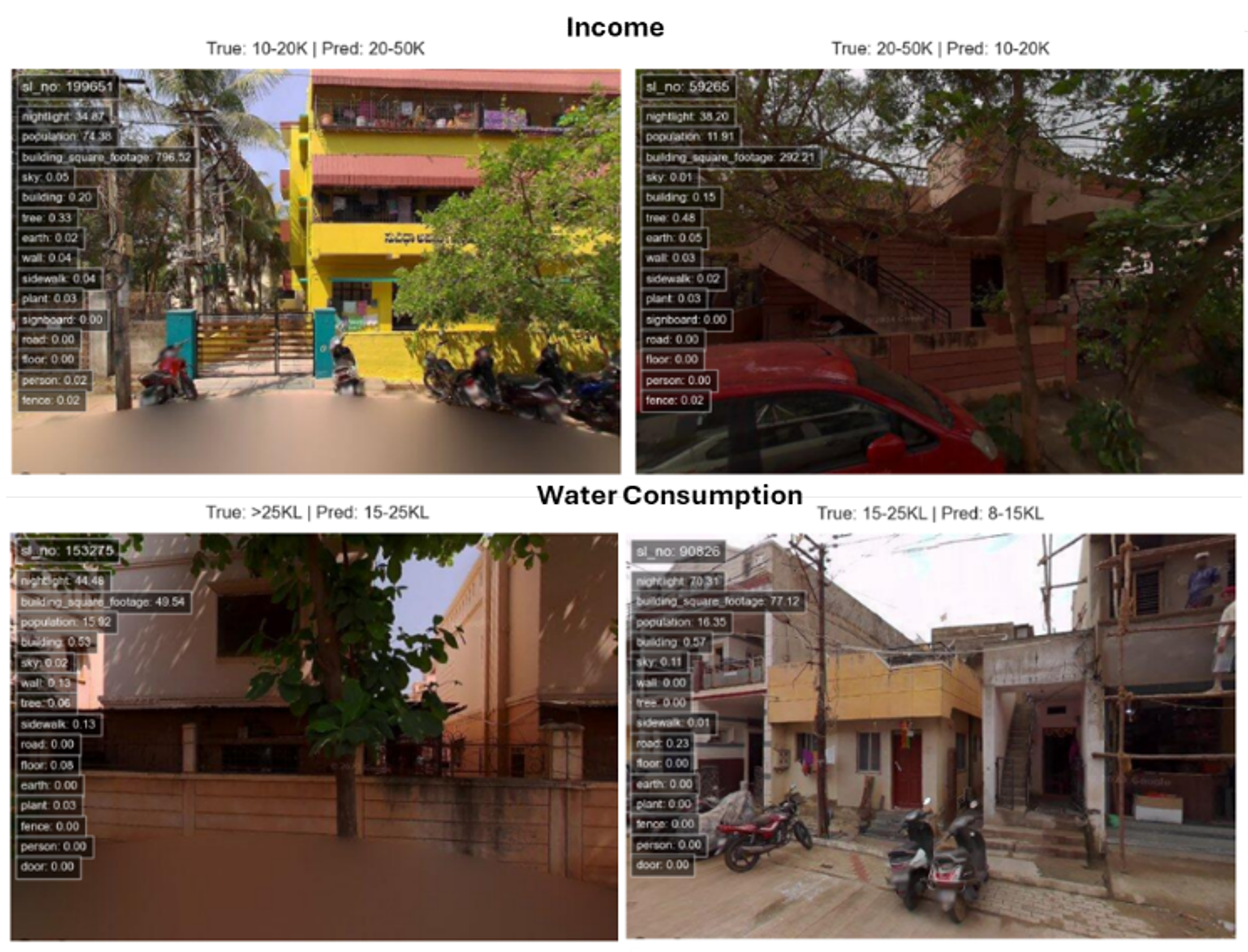}
    \caption{\textbf{Illustrative examples of misclassified cases based on street view segmentation features.} 
Top panels: income predictions, where a household in the Rs.~10--20K bracket is predicted as Rs.~20--50K, and a household in the Rs.~20--50K bracket is predicted as Rs.~10--20K. 
Bottom panels: water consumption predictions, where a household with $>$25 KL consumption is predicted as 15--25 KL, and a household with 15--25 KL consumption is predicted as 8--15 KL.}
    \label{fig7}
\end{figure}

A careful review of misclassified cases using Google Street View segmentation features reveals several recurring sources of error. For income prediction, misclassifications tend to occur in areas where visual cues are ambiguous or contextually misleading—for example, modest homes with well-maintained facades, densely packed neighborhoods with a mix of building types, or scenes partially obscured by vegetation or roadside clutter. In such settings, typical visual signals of socio-economic status become less distinct, particularly in transitional or mixed-use zones. In the case of water consumption prediction, segmentation-based features appear even less informative. Misclassified samples frequently lack visible indicators of domestic infrastructure, such as water storage tanks, open yards, or distinguishable roof features. Moreover, dense vegetation or occluded views often dominate these images, further diminishing the model’s ability to extract relevant information. Illustrative examples of such misclassified cases are shown in Fig.~\ref{fig7}, highlighting the limitations of using street view segmentation alone as a proxy for socio-economic or behavioral characteristics, especially in environments where such attributes are not reliably expressed through exterior visual features.

\section*{Discussion and Conclusion}

 This study demonstrates the promise and limitations of using remote sensing and street-level imagery to predict household-level   water consumption in rapidly urbanizing regions in comparison with household level subjective income. Among the image-based features, satellite-derived nightlight intensity and building footprint showed strong associations with both water consumption and income levels. Models that incorporated embeddings from GSV images captured relevant visual cues, but their performance plateaued in areas with ambiguous or transitional built environments.

Misclassification analysis revealed systematic challenges in distinguishing middle-income and mid-range-consumption households, particularly in mixed-use or visually heterogeneous neighborhoods. Visual inspection of GSV images indicated that infrastructure features critical to income or water usage—such as plot size, roofing materials, or presence of vegetation—can be visually subtle, occluded, or inconsistent across contexts. These limitations underscore the difficulty of inferring socio-economic status purely from visual proxies without contextual or geographic grounding.

Moving forward, several extensions could improve model performance and generalizability. First, incorporating spatial clustering or neighborhood-level priors—such as ward-level averages or localized covariates—may help to reduce ambiguities in samples from belonging to the borders of income or consumption groups. Second, semantic segmentation can be further refined to isolate structures (e.g., rooftops, yards, water tanks) that are more functionally tied to the outcomes of interest. Third, expanding the dataset to include additional cities with diverse urban morphologies would allow for rigorous out-of-sample testing and domain transfer evaluation. Finally, multimodal integration—combining imagery with household survey data, utility records, or administrative boundaries—offers a promising direction for building more accurate and policy-relevant models. Overall, while visual data provides a scalable avenue for socio-economic inference, they should be interpreted in conjunction with spatial context and domain-specific knowledge, especially in data-sparse or rapidly evolving urban environments.

\section*{Acknowledgments}

This study was conducted as part of the Impact Evaluation of the Karnataka Urban Water Supply Modernization Project (KUWSMP), implemented by the Government of Karnataka with financial and technical support from the World Bank. The authors gratefully acknowledge the collaboration of the Karnataka Urban Infrastructure Development and Finance Corporation (KUIDFC), the Project Management Unit in Hubballi-Dharwad, and the field survey teams for their contributions to data collection and validation. The findings, interpretations, and conclusions expressed in this paper are those of the authors and do not necessarily reflect the views of the World Bank, its Executive Directors, or the governments they represent.

%\nolinenumbers

% Please compile your BiBTeX database using the "plos2025.bst" BibTeX style.
% This file is part of the current package.
% A sample BibTeX file is also included as "plos_bibtex_sample.bib".
%
% or
%
% Type in your references following Vancouver style and reference formatting instructions
% available at https://journals.plos.org/plosone/s/submission-guidelines#loc-references
% \begin{thebibliography}{}
% Text
% \end{thebibliography}

\bibliography{plos_bibtex_sample}

\end{document}